\pdfoutput=1
\documentclass{amia}
\usepackage{graphicx}
\usepackage[labelfont=bf]{caption}
\usepackage[superscript,nomove]{cite}
\usepackage{color}
\usepackage{amsmath}
\usepackage{ulem} 
\usepackage[dvipsnames]{xcolor}
\usepackage{bm}

 \renewcommand{\sout}[1]{\iffalse {#1} \fi}

\begin{document}

\title{Retrofitting Vector Representations of Adverse Event Reporting Data to Structured Knowledge to Improve Pharmacovigilance Signal Detection}
\author{Xiruo Ding, MS$^{1}$, Trevor Cohen, MBChB, PhD$^{1}$}

\institutes{
    $^1$University of Washington, Seattle, WA, USA\\
}

\maketitle

\noindent{\bf Abstract}

\textit{Adverse drug events (ADE) are prevalent and costly. Clinical trials are constrained in their ability to identify potential ADEs, motivating the development of spontaneous reporting systems for post-market surveillance. Statistical methods provide a convenient way to detect signals from these reports but have limitations in leveraging relationships between drugs and ADEs given their discrete count-based nature. A previously proposed method, \texttt{aer2vec}, generates distributed vector representations of ADE report entities that capture patterns of similarity but cannot utilize lexical knowledge. We address this limitation by retrofitting \texttt{aer2vec} drug embeddings to knowledge from RxNorm and developing a novel retrofitting variant using vector rescaling to preserve magnitude. When evaluated in the context of a pharmacovigilance signal detection task, \texttt{aer2vec} with \texttt{retrofitting} consistently outperforms disproportionality metrics when trained on minimally preprocessed data. \texttt{Retrofitting} with rescaling results in further improvements in the larger and more challenging of two pharmacovigilance reference sets used for evaluation.}

\section*{Introduction}
Adverse events (AE) were first defined as ``injuries caused by medical management, and of the subgroup of such injuries that resulted from negligent or substandard care''\cite{brenann1991}. As a subgroup of AE, adverse drug events (ADE) are thus defined as injuries due to medication. ADEs are common, costly, and sometimes preventable. Kaushal et al reported an ADE rate of 10\% (10 potential ADEs per 100 admissions) in two pediatric care clinics \cite{kaushal2001}. A report entitled \textit{Preventing Medication Errors}, by the Institute of Medicine, estimated a cost of \$1,983 per preventable ADE, with national annual cost of \$887 million, while claiming that at least a quarter of all ADEs are preventable \cite{PreventingMedicationErrors}. A challenge for prevention is that many adverse effects of drugs are not known before their release to market, in part because clinical trials are not sufficiently large or inclusive to account for the variety of human responses to a given pharmaceutical agent \cite{pharmacoBook}.

To address this issue, the World Health Organization (WHO) established the WHO Global Monitoring Programme in 1968, with 10 pilot countries. However, the international community did not recognize the importance of sharing ADE information until 1982 when benoxaprofen was reported to cause 12 deaths in the two years after its initial release in the US \cite{oraflex}.  Subsequently, the community made changes to adopt better epidemiological techniques and information systems to share information. With national and international organizations, such as the United States Food and Drug Administration (FDA), the WHO and the European Medicines Agency (EMA), countries are developing their own reporting system infrastructures. The spontaneous reporting system in the U.S. is managed by the FDA, with voluntary adverse drug reaction case reports sent directly to the FDA or drug manufacturers by healthcare professionals and consumers, and retained in the FDA Adverse Event Reporting System (FAERS) database.

In order to detect ADEs from FAERS data, several statistical analyses have been proposed and are in current use, including the Proportional Reporting Ratio (PRR) \cite{prr2001}, Reporting Odds Ratio (ROR) \cite{ror2004}, and Multi-Item Gamma Poisson Shrinker (MGPS) \cite{mgps}. The PRR and ROR are derived from report-level statistics with straightforward calculations, and the MGPS is conceptually similar to the PRR with additional Bayesian shrinkage and stratification, which are useful with limited data \cite{mgps}. These traditional statistical methods are limited by their inability to draw relations between similar drugs or between similar adverse events. This is analogous to a widely recognized problem with Natural Language Processing (NLP) techniques based on discrete counts of words, which fail to recognize useful relationships between them. There are two sorts of generalization that could improve performance of discrete count-based methods: (1) generalization based on patterns of distribution (entities with similar distributions are similarly represented), as exemplified by methods of distributional semantics such as skip-gram with negative sampling (SGNS - a component of the popular \texttt{word2vec} package)\cite{SGNS} and GloVe \cite{glove}; and (2) generalization based on structured knowledge available in a lexicon. The second approach can be accomplished by combining methods of distributional semantics with \texttt{retrofitting}, a method that uses structured knowledge to enhance distributionally-derived representations \cite{retrofitting2014}. In the first case, the model learns to recognize patterns in the data, but doesn't draw on knowledge beyond the training corpus. In the second, the model should recognize both those relations implicit in the data and those explicitly defined in a relevant knowledge resource.

In this paper, we first use a previously proposed algorithm, \texttt{aer2vec}  \cite{aer2vec2019}, which uses a similar architecture to SGNS with variants that consider either drugs or ADEs as inputs at FAERS report level (rather than using a sliding window based unit of context) to build distributed representations of drugs and ADEs. Unlike prior work with \texttt{aer2vec} \cite{aer2vec2019}, we apply the algorithm to FAERS data as is - without additional normalization and standardization. Next, we leverage the \texttt{retrofitting} algorithm of Faraqui et al \cite{retrofitting2014} to combine distributional information of drugs with lexical information from RxNorm, a U.S. National Library of Medicine project under the parent project of the Unified Medical Language System (UMLS) \cite{bodenreider2004unified}. Two key distinctions between our use of these methods and their original conception are that we are using them to represent reporting data rather than language, and using the resulting vector representations to recapitulate a probabilistically-motivated training objective (e.g. $P(drug|ADE)$) rather than comparing input weight vectors directly (as is typical in word embedding applications). Our hypothesis is that retrofitting \texttt{aer2vec} vectors will result in improvements in performance on pharmacovigilance signal detection tasks.

\section*{Methods}
The underlying idea of our methods is to use \texttt{aer2vec} to train a set of drug embeddings, and then modify these to conform with lexical information derived from RxNorm using a \texttt{retrofitting} algorithm that was originally developed to enrich word embeddings with information from different lexical sources \cite{retrofitting2014}. The training set for drug and ADE embeddings is derived from FAERS. As with unmodified \texttt{aer2vec} representations, the probabilistically-motivated training objective can be recapitulated to estimate $P(drug|ADE)$. The resulting models are then evaluated on two widely-used pharmacovigilance reference sets, one from the Observational Medical Outcomes Partnership (OMOP) \cite{omop}, the other from the EU-ADR (Exploring and Understanding Adverse Drug Reactions) Project \cite{euset}.

\textbf{\underline{\textsc{FDA Adverse Event Reporting System Data}}} The FDA actively maintains the FAERS post-market surveillance system. The system has been active since 1969, and experienced a major redesign in 1997, when the Medical Dictionary for Regulatory Activities (MedDRA) was used to harmonize medical terminologies  \cite{faers}. Both drugs and ADEs in FAERS were mapped to this standardized dictionary. However, not all publicly available fields in the database were mapped. For example, one important field for our study, DRUGNAME, was mapped to active ingredients (a field named PROD\_AI) only after the third quarter of 2014. To test model performance using raw data with minimal preprocessing and maximize data use, the DRUGNAME field was used in the current study. Since the source of FAERS data is spontaneous reports, this  field is not very well standardized. This required some minimal processing: (1) all entries were changed into lower case; (2) spaces were replaced with underscores; (3) trailing special characters were removed. ADEs have been mapped to canonical terms in a medical terminology (a MedDRA ``Preferred Term'') since the inception of the system, so this field was used directly in the study without any modification.

FAERS data include reporter designations of suspicion for drugs a patient was taking when an ADE was observed. We used 13,636,783 publicly available records from 2004 to 2019, all but 134 of which have at least one designated primary suspect (PS). There are other reported roles of drugs besides PS in the reports, such as SS (secondary suspect), C (concomitant), and I (interacting). The main focus of the study is on both a set of all roles, and a set including primary suspect drugs (PS) only. Restricting data to drugs designated with the PS identifier was shown to  improve model performance when evaluated using standard reference sets \cite{aer2vec2019}. Consequently, we constructed two sets of report data: FULL, which includes all report-level drug/ADE co-occurrence events, and PS which includes only those drugs from a report that have been designated as primary suspects. These sets were used as training data for two independent \texttt{aer2vec} models.

\textbf{\underline{\textsc{\texttt{aer2vec}}}} The \texttt{aer2vec} algorithm\cite{aer2vec2019} adapts the SGNS algorithm of Mikolov et al \cite{word2vec1, word2vec2} to represent report-level drug/ADE co-occurrence events. Retaining SGNS as a basic training framework, \texttt{aer2vec} introduces two neural architectures, \texttt{aer2vec+} and \texttt{aer2vec-}, to predict $P(drug|ADE)$ and $P(ADE|drug)$ respectively. SGNS is trained to predict the probability of observing each context term in a sliding window surrounding an observed term, which is moved through a training corpus. In contrast, \texttt{aer2vec} models context at the report level where co-occurrence events are observed, providing data for training of the relevant neural network architecture. As similar representations will be learned for entities occurring in similar contexts, connections between similar drugs (such as drugs resulting in the same ADEs) or similar ADEs (such as those resulting from the same drugs) can be inferred. As \texttt{aer2vec+} performed better on signal detection tasks\cite{aer2vec2019}, we employed it for the current work. The FULL and PS subsets of FAERS data were used to train ADE and drug embeddings using  \texttt{aer2vec+}, as implemented in the open source \texttt{Semantic Vectors}\footnote{https://github.com/semanticvectors/semanticvectors} package \cite{semantic2008, semantic2010}, following the accompanying documentation\footnote{https://github.com/treversec/aer2vec}. In accordance with the best-performing models described in the original work\cite{aer2vec2019}, vectors of dimension of 100 for every drug (DRUGNAME) and ADE (PT) were trained over ten iterations on each training set (without subsampling of frequently occurring entities). The resulting embeddings provided initial information on drug and ADE distributions in the reports. Since \texttt{aer2vec} training depends upon randomized order of presentation of the data and stochastic initialization of weight vectors, we generated ten editions of each of the two embedding spaces, in order to assess robustness of performance.

\textbf{\underline{\textsc{RxNorm Relations}}} RxNorm is a project that started in 2001, with the goal of modeling clinical drugs and their relationships under the Unified Medical Language System (UMLS) \cite{rxnormHistory}. RxNorm provides extensive information on clinical drugs,  links across drug vocabularies, and drug interactions \cite{rxnormdoc}. RxNorm is organized by concepts (denoted as RxCUIs), adopting the structure used in the UMLS semantic network for a consistent categorization \cite{umlsRef}. Each concept then links to many related terms. For example, the concept of ``rofecoxib'' (RxCUI = 232158) is linked to related terms where it is the ingredient,  ``rofecoxib'' and ``pridinol/rofecoxib'', and its pharmaceutical preparations ``rofecoxib 12.5 MG'', ``rofecoxib 2.5 MG/ML'' in the RxNorm database. For each concept (RxCUI), one or several atoms (RxAUI) are categorized under it. For the same example of ``rofecoxib'', it has atoms of ``rofecoxib'', ``4-[4-(methylsulfonyl)phenyl]-3-phenyl-2(5H)-furanone'', ``3-phenyl-4-[4-(methylsulfonyl)phenyl]-2(5H)-furanone'' (with the latter two being chemical designations of the drug), amongst others. These terms could be used in mapping to the DRUGNAME field of the FAERS dataset.

RxNorm is released on a monthly basis. For this study, the version released on January 6th, 2020 was used, including 6,574,327 relational records in total. Seven RxNorm relation types (RELs) are provided: RO, SY, RB, RN, SIB, CHD, and PAR. In previous work evaluating the effect of retrofitting biomedical concept embeddings on the agreement between their pairwise similarities and human judgment,  the combination of relationships RN (``narrower'' relationship) and RO (other relationship) improved performance \cite{retrofittingUMLS2017}. In our study, SY (synonymy) was also included as a lexical relationship with RN and RO, thus covering most instantiated RxNorm relations. Since RxNorm relations are organized by RxCUIs, all associated atomic concepts were assumed to inherit relations from their parent RxCUI. The resulting lexicon included 668,412 entries, with each RxCUI or RxAUI being an entry to retain the most granular information.

\textbf{\underline{\textsc{\texttt{retrofitting}}}} \texttt{retrofitting} is a graph-based technique developed by Faruqui et al in order to integrate vector space representations (word embeddings) with semantic lexicons \cite{retrofitting2014}. The idea behind this method is to update the target's vector representation by bringing it closer to those of its lexical neighbors, thus integrating information from them. This can be summarized in the following cost function to be minimized \cite{retrofitting2014}:
\begin{align}
	\Psi (A) = \sum_{i=1}^{n}\left[\alpha_i ||q_i - \hat{q}_i ||^2 + \sum_{(i,j) \in E}\beta_{ij} ||q_i -q_j ||^2   \right]
\end{align}
where $E$ is the set of semantic relations of interest, $q_i$ the retrofitted representation for concept $i$ (target), $\hat{q}_i$ the original vector representation for this concept, $q_j$ a neighbor of $q_i$ in $E$, and model parameters $\alpha_i$ the weight for original vector representations, and $\beta_{ij}$ the weight for lexical information.

The method is able to retrofit vector representations generated by any algorithm. For the current work drug vectors from \texttt{aer2vec+} trained on the FULL and PS sets were used. The lexical information is from the RxNorm lexicons described in the  previous section covering relations RN, RO and SY. \texttt{Retrofitting} was accomplished using the implementation provided by Faraqui and his colleagues\footnote{https://github.com/mfaruqui/retrofitting} with minor modifications to avoid discarding terms with certain characters. While Faruqui et al reported results with fixed hyperparameters after setting $\alpha$ = $\beta$ = 0.5, we experimented with adjusting their values to evaluate the effect on performance of balancing distributional information against lexical information. This was done by changing the value of $\beta$ from 0 to 1 by increments of 0.1 while maintaining $\alpha + \beta = 1$.

\textbf{\underline{\textsc{Rescaling}}} In the original implementation of \texttt{retrofitting}, word vector representations are normalized before proceeding. However, with \texttt{aer2vec+}, $P(drug|ADE)$ is estimated as $\sigma(\overrightarrow{\bm{ADE}} \cdot \overrightarrow{\bm{drug}})$ with $\overrightarrow{\bm{ADE}}$ and $\overrightarrow{\bm{drug}}$ representing the input and output weights of the neural network, respectively. This estimate is sensitive to the magnitude of the weight vectors concerned, which is discarded upon normalization. To preserve this valuable information, a novel rescaled version of \texttt{retrofitting} was developed, by using the lengths of the original \texttt{aer2vec} representations to determine the lengths of their retrofitted counterparts:
\begin{align}
	\bm{vrr_i} = \frac{|| \bm{v_i } ||_2}{|| \bm{vr_i} ||_2} \times \bm{vr_i}
\end{align}
where for word $i$, $\bm{vrr_i}$ is rescaled retrofitted vector representation, $\bm{vr_i}$ is retrofitted vector representation, $\bm{v_i}$ is vector representation from \texttt{aer2vec}. The euclidean norm is used for normalization. Similarly, these rescaled vectors were evaluated under all pairs of $\alpha$ and $\beta$ hyperparameter values.

\textbf{\underline{\textsc{Baseline Models}}} Disproportionality metrics, including the Proportional Reporting Ratio (PRR) and the Reporting Odds Ratio (ROR), are widely-used pharmacovigilance methods to detect signals in spontaneous reporting system data, such as in FAERS \cite{fdaPRR} and the UK Yellow Card database\cite{ukPRR2001}. Both PRR and ROR are derived from the $2\times2$ table for a specific drug-ADE pair:

\begin{table}[H]
	\centering
	\caption{Disproportionality Metrics}
	\begin{tabular}{|l|c|c|}
		\hline
		& \textbf{ADE of Interest}  & \textbf{Other ADEs} \\ \hline
		Drug of Interest  & a & b  \\ \hline
		Other Drugs   & c  & d \\ \hline
		
	\end{tabular}
	\label{table:dm}
\end{table}

Given Table ~\ref{table:dm} where $a,b,c,d$ are all counts for co-occurrences, PRR and ROR can be expressed as:
\begin{align}
PRR &= \frac{a}{a+b}/\frac{c}{c+d} \\
ROR &= \frac{a}{b}/\frac{c}{d}
\end{align}

The strengths of the PRR are its convenience to calculate, and that it is not affected by bias from varying reporting frequencies \big{(}unlike methods such as reporting rates (\# of reports divided by \# of prescriptions)\big{)}, because the numerator estimates  the proportion of reports for a drug of interest that concern a particular ADE. However, one limitation of the PRR is that a strong signal from a particular drug will reduce the magnitude of signals for other drug-ADE pairs \cite{prr2001}. The ROR method makes predictions based on estimation of odds ratios, and this can be used to estimate relative risk if the database is modified to simulate a case-control study by removing related drug-ADE pairs to create a ``randomized'' control set \cite{ror2004}. However, from the formula, it is apparent that the ROR will not compute when any of $b$, $c$ or $d$ are 0. Practically, PRR and ROR are similar and give comparable results when used for signal detection, in terms of sensitivity and specificity \cite{Egberts2002}. A common concern for these  methods is that they treat drugs and ADEs as discrete entities, and as such cannot generalize from data concerning related drugs (such as class effects) or ADEs (such as organ system effects). It is these limitations that our methods aim to address.

\begin{figure}[t]
	\centering
	\captionsetup{justification=centering}
	\includegraphics[scale=0.7]{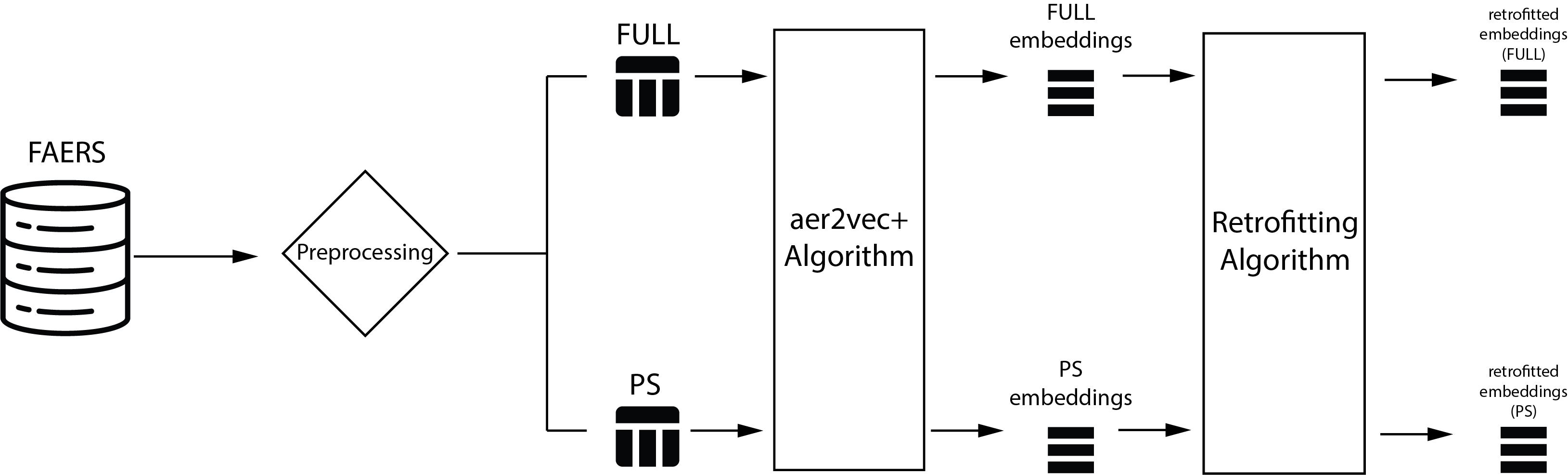}
	\caption{Pipeline to generate embeddings from data subsets, with and without retrofitting.}
	\label{fig:workflow}
\end{figure}

\textbf{\underline{\textsc{Evaluation Sets}}} To evaluate model performance, two reference sets were used in the study, one from the Observational Medical Outcomes Partnership (referred to as the OMOP set), the other one from the EU-ADR (Exploring and Understanding Adverse Drug Reactions) Project (referred to as the EU set). The OMOP set was developed through systematic literature review and natural language processing of structured product labels to help establish classifications of positive and negative drug-ADE pairs \cite{omop}. To limit the scope of research, the group defined four main outcomes: acute liver injury, acute kidney injury, acute myocardial infarction, and upper gastrointestinal bleeding. For the four outcomes, 165 positive controls and 234 negative controls were identified. The EU set was built in a similar way by integrating literature, drug product labels, and expert opinion, however unlike the OMOP set spontaneous report data from the World Health Organization was also used, and they were key to generating the pool of negative controls from which this aspect of the set was derived. Ten adverse events (bullous eruptions, acute renal failure, anaphylactic shock, acute myocardial infarction, rhabdomyolysis, aplastic anaemia/pancytopenia, neutropenia/agranulocytosis, cardiac valve fibrosis, acute liver injury, and upper gastrointestinal bleeding) were used as outcomes, and 94 drug-ADE pairs accounting for 44 positive controls and 50 negative controls across these outcomes were then identified \cite{euset}. 
For these two evaluation sets, performance was evaluated using the Area Under the Receiver Operating Characteristic Curve (henceforth AUC). The five methods, \texttt{aer2vec}, \texttt{aer2vec} with \texttt{retrofitting} (with and without rescaling), and two baseline models, were each applied to two separate data subsets: FULL and PS, and evaluated using the two reference sets, OMOP and EU-ADR. With all \texttt{aer2vec} variants, $P(drug|ADE)$ was estimated as $\sigma(\overrightarrow{\bm{ADE}} \cdot \overrightarrow{\bm{drug}})$, and used to score the drug/ADE pairs in the sets. The entire workflow is illustrated in Figure \ref{fig:workflow}.

\section*{Results}

\textbf{\underline{\textsc{aer2vec+}}} When training \texttt{aer2vec+} on the FULL and PS sets, and evaluating on the EU and OMOP reference sets,  we obtain the AUC results shown in Table \ref{table:aer2vec}, which also includes baseline PRR and ROR results.

\begin{table}[H]
	\centering
	\caption{\texttt{aer2vec+} performance (AUC) vs PRR/ROR}
	\begin{tabular}{c|c|c|c|c}
		\hline
		& \multicolumn{2}{c|}{\textbf{OMOP}} & \multicolumn{2}{c}{\textbf{EU}} \\
		\hline
		& FULL        & PS       & FULL         & PS        \\
		\hline
		aer2vec+ 	&      $0.72 \pm 0.011$      &     $0.84 \pm 0.005$     &       $0.80 \pm 0.020$         &       $0.96 \pm 0.005$    \\
		PRR      	&0.64            &   0.73       &     0.83         &   0.89        \\
		ROR      	&   0.64          &  0.73       &     0.83         &   0.89       \\
		\hline
	\end{tabular}
	\label{table:aer2vec}
\end{table}

When evaluating the method on the OMOP set (Table \ref{table:aer2vec} left),  \texttt{aer2vec+} performs significantly better than the PRR and ROR in both cases (PS and FULL). We find a similar pattern for the EU reference set (Table \ref{table:aer2vec} right), with the exception that the \texttt{aer2vec+} algorithm performs slightly worse  than the PRR or ROR statistics when trained on the FULL set. The AUCs of \texttt{aer2vec+} rise significantly from FULL to PS set in both evaluation sets, which is also true for the baseline models of PRR and ROR. Both findings are consistent with previous work using a normalized standardized set of FAERS data \cite{aer2vec2019}. The overall performance of all three methods on the EU set is better than their performance on the OMOP set. Moreover, with largest training set and smallest test set (FULL - EU), \texttt{aer2vec+} has the highest standard deviation of 0.02, as compared with that of 0.005 with the PS - OMOP set combination.

\begin{figure}[t]
	\centering
	\captionsetup{justification=centering}
	\includegraphics[scale=0.98]{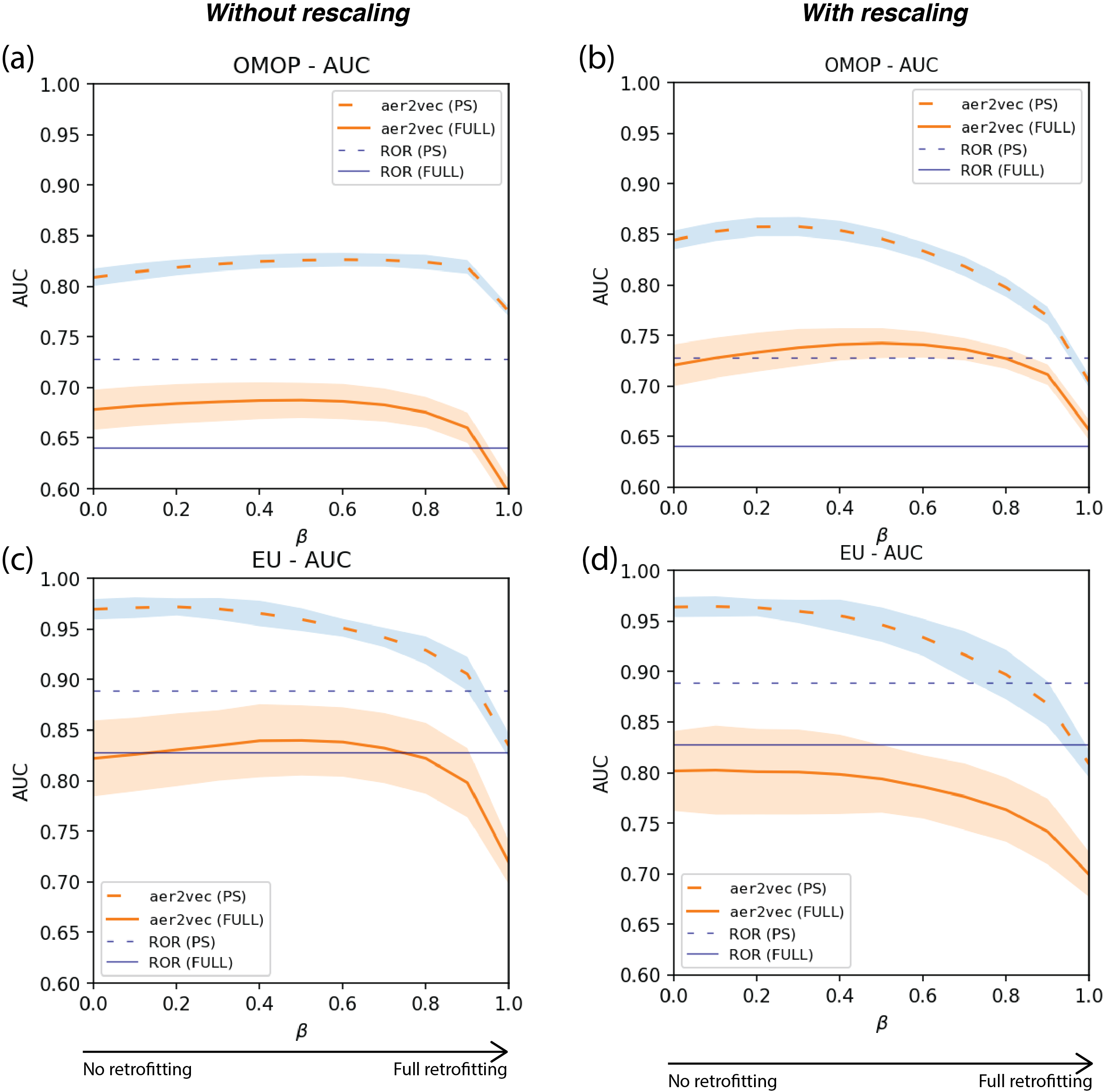}
	\caption{Effects of \texttt{retrofitting} on AUCs for FULL and PS set (10 iterations).\\ 
		(a) \texttt{retrofitting} without rescaling on OMOP set (b) \texttt{retrofitting} with rescaling on OMOP set \\
			(c) \texttt{retrofitting} without rescaling on EU set (d) \texttt{retrofitting} with rescaling on EU set.}
	\label{fig:retrofit_beta}
\end{figure}

\textbf{\underline{\textsc{Retrofitting of drug embeddings}}} Results with \texttt{retrofitting} are shown in Figure ~\ref{fig:retrofit_beta} (left). Given equation (1), $\beta$ (represented by the $x$ axis) with value of 0 means there is no retrofitting involved, and all resulting vectors are from the original \texttt{aer2vec+} models (albeit with normalization when not rescaled). $\beta$ with value of 1 means all vectors come from fully retrofitted models, where the resulting vectors are updated without the constraint that they should remain similar to their \texttt{aer2vec}-derived starting points, as determined by lexical information from RxNorm. The AUCs of models trained on the PS set are always higher than those from the FULL set, whatever the value of $\beta$ is ($p < 0.01$,  unpaired t-test). This is consistent with the findings described in the previous section where no \texttt{retrofitting} was used at all, however we note that the normalization step of the \texttt{retrofitting} method results in lower baseline ($\beta=0$) performance with the OMOP set ($p<0.05$,  unpaired t-test), and a slightly higher baseline performance with the EU set ($p<0.05$,  unpaired t-test). 

When the \texttt{retrofitting} algorithm is applied (Figure ~\ref{fig:retrofit_beta} (a) (c)), the model performance increases as the weight of retrofitting increases, but declines after a point with worst performance when lexical information is permitted to predominate ($\beta=1$). This is true for both FULL and PS sets when evaluated on both EU and OMOP reference sets. However, the degree of improvement differs across these reference sets. With the OMOP set, \texttt{aer2vec+} has a larger increase in AUC when trained on the PS set. Conversely, when evaluated on the EU set the algorithm trained on the FULL set benefits more. The difference from these evaluation sets also shows as the $\beta$ values for algorithm to achieve its best performance: in the EU set, it is $\beta = 0.5$ (FULL set) and $\beta = 0.2$ (PS set); in OMOP set, it is $\beta=0.5$ (FULL set) and $\beta=0.6$ (PS set). These results show that information in lexicons can be utilized to improve \texttt{aer2vec}'s performance in pharmacovigilance signal detection. 

The rescaling results are shown in Figure ~\ref{fig:retrofit_beta} (right). When evaluated on the OMOP set ((a) and (b) in Figure ~\ref{fig:retrofit_beta}), both models (PS and FULL training sets) benefit from rescaling, with an overall gain in AUC of 0.05 on the FULL set and 0.04 on the PS set. Moreover, the retrofitting effect is more obvious when vectors are rescaled: when $\beta$ increases, AUC also increases and reaches its maximum when $\beta=0.5$ for the FULL set. A similar pattern is also observed with the PS set, but with a maximum AUC at a lower $\beta$ (0.3), and the AUC dropping more quickly after the maximum point. After applying rescaling, the results on OMOP set are shown in Table ~\ref{table:omop_beta_rescale}. Since the results of PRR and ROR are very close, only ROR is shown in tables. Compared with \texttt{aer2vec} without rescaling, we observe an overall increase in AUCs, with different $\beta$s used to achieve maximal performance. These are the best results obtained with the \texttt{aer2vec} method on the OMOP evaluation set in the current work.
\begin{table}[t]
	\centering
	\caption{\texttt{Retrofitting} results (part) on the OMOP set with varying $\beta$. Best results in \textbf{boldface} with ROR and \texttt{aer2vec} ($\beta$ = 0.5) as points of comparison.}
	\begin{tabular}{c|c c|c c}
		\hline
		 & \multicolumn{2}{c|}{Without rescaling} &  \multicolumn{2}{c}{With rescaling} \\
		\textbf{$\beta$}  &  \textbf{FULL}      & \textbf{PS}      & \textbf{FULL}             & \textbf{PS}         \\
		\hline
		0    & $0.678\pm 0.010$ & $0.809\pm 0.004$ & $0.721\pm 0.011$ & $0.844\pm 0.005$ \\
		0.3  & $0.686 \pm 0.010$  & $0.822\pm 0.004$ & $0.738\pm 0.009$ & $\mathbf{0.858\pm 0.005}$ \\
		0.5  & $\mathbf{0.687\pm 0.009}$ & $0.826\pm 0.004$ & $\mathbf{0.742\pm 0.008}$ & $0.846\pm 0.005$ \\
		0.6  & $0.686\pm 0.009$ & $\mathbf{0.826\pm 0.003}$ & $0.741\pm 0.007$ & $0.834\pm 0.004$ \\
		1.0    & $0.596\pm 0.007$ & $0.775\pm 0.003$ & $0.656\pm 0.005$ & $0.705\pm 0.003$ \\
		\hline
		ROR   & $0.641$ & $0.728$ & $0.641$ & $0.728$\\
		\hline
	\end{tabular}
	\label{table:omop_beta_rescale}
\end{table}
In contrast, when comparing (c) and (d) in Figure ~\ref{fig:retrofit_beta}, performance on the EU set does not benefit from rescaling, with slight decreases with both training sets. Furthermore, when evaluated on this set, the normalized and rescaled vector representations have similar retrofitting effects, with slight increases in AUC as the influence of \texttt{retrofitting} is  initially increased. Table \ref{table:eu_beta_rescale} shows partial EU set results. Rescaled \texttt{retrofitting} performs best at $\beta=0.1$, but \texttt{retrofitting} without rescaling performs best overall. \vspace{5pt}

\begin{table}[H]
	\centering
	\caption{\texttt{Retrofitting} results (part) on the EU set with varying $\beta$. Best results in \textbf{boldface} with ROR and \texttt{aer2vec} ($\beta$ = 0.5) as points of comparison.}
	\begin{tabular}{c|c c|c c}
		\hline
		& \multicolumn{2}{c|}{Without rescaling} &  \multicolumn{2}{c}{With rescaling} \\
		\textbf{$\beta$}  &  \textbf{FULL}      & \textbf{PS}      & \textbf{FULL}             & \textbf{PS}         \\
		\hline
		0      & $0.822\pm 0.019$ & $0.969\pm 0.005$ & $0.802\pm 0.020$ & $0.964\pm 0.005$\\
		0.1   & $0.826 \pm 0.019$  & $0.971\pm 0.005$ & $\mathbf{0.803\pm 0.023}$ & $\mathbf{0.964\pm 0.005}$ \\
		0.2   & $0.831\pm 0.018$ & $\mathbf{0.972\pm 0.004}$ & $0.801\pm 0.022$ & $0.963\pm 0.004$ \\
		0.5   & $\mathbf{0.840\pm 0.018}$ & $0.959\pm 0.006$ & $0.794\pm 0.017$ & $0.946\pm 0.009$ \\
		1.0    & $0.719\pm 0.011$ & $0.834\pm 0.006$ & $0.699\pm 0.011$ & $0.809\pm 0.007$\\
		\hline
		ROR   & $0.828$ & $0.889$ & $0.828$ & $0.889$\\
		\hline
	\end{tabular}
	\label{table:eu_beta_rescale}
\end{table}

\section*{Discussion}
In this paper, we evaluated the utility of combining two methods, \texttt{aer2vec} and \texttt{retrofitting}. We hypothesized that this combination would improve pharmacovigilance performance with \texttt{aer2vec} when trained on a minimally preprocessed FAERS dataset, without normalization or standardization. The \texttt{aer2vec} method was developed to model empirical ADE reports for pharmacovigilance signal detection, and showed its utility when trained on standardized FAERS data \cite{aer2vec2019}. Its application to minimally preprocessed FAERS data has not been evaluated previously. The \texttt{retrofitting} method was originally used to refine semantic vector space representations of words\cite{retrofitting2014}, and its application to retrofit data-derived concept embeddings for the purpose of predictive modeling is a novel aspect of the current work, as is the \textit{rescaling} modification of this method to preserve information encoded by vector magnitude. Evaluations on both EU and OMOP sets show improvements with \texttt{retrofitting} over \texttt{aer2vec} as the retrofitting effect ($\beta$) increases, but the optimal $\beta$  differs from set to set, as does the influence of rescaling. On the larger and more challenging OMOP set, the combination of these two methods (\texttt{retrofitting} with rescaling) produces an improvement over baseline models, the PRR and ROR. On the EU set, \texttt{retrofitting} alone leads to best performance, comfortably outperforming these disproportionality metrics, which are in current use for pharmacovigilance signal detection. This illustrates their capability as a means of mining spontaneous ADE reports, and highlights the potential utility of combining empirically-derived embeddings with lexical information. Lexical information from RxNorm serves as a complement to distributional information learned from the data directly, which in our case includes data from spontaneous reports but excludes drug classes, relations, and hierarchy, just as most distributional semantics algorithms don't take lexical knowledge beyond their training corpus into account.

The results also illustrate the feasibility and stability of \texttt{aer2vec} when trained on a minimally preprocessed FAERS dataset, without standardization or normalization. The \texttt{aer2vec+} model trained on raw data performs relatively well. Portanova et al reported best AUCs of 0.86 and 0.96 on the EU set when training on the FULL and PS sets respectively, and 0.76 and 0.88 on the OMOP set when training on the FULL set and the PS set \cite{aer2vec2019}. The performance of the same algorithm trained on the raw FAERS dataset (shown where $\beta=0$ with rescaling in Table \ref{table:omop_beta_rescale} and Table \ref{table:eu_beta_rescale}) was not better than that on the standardized set. However, combining \texttt{aer2vec} and \texttt{retrofitting} (with and without rescaling) achieved the best results on these two reference sets in the current experiments, surpassing results with the standardized set in one instance. Our models achieved best AUCs of 0.84 ($std=0.018$) and 0.97 ($std=0.004$) on the EU set ( \texttt{retrofitting} without rescaling)  when trained on the FULL and PS sets respectively, and 0.74 ($std=0.008$) and 0.86 ($std=0.005$) on the OMOP set (\texttt{retrofitting} with rescaling) when trained on the FULL and PS sets respectively, which suggests that \texttt{retrofitting} can to some degree compensate for the lack of standardization and normalization with raw FAERS data. However, these results are not strictly comparable because Portanova et al's model was trained using data from up to the 2015 release only. The availability of more current data than that which is available in standardized form shows one advantage of utilizing minimally preprocessed data where little effort is required for harmonizing and mapping of terms. Another advantage is that the computational work required to perform this mapping would need to be repeated for the entire data set if any change in relational structure occurred, whereas the \texttt{retrofitting} procedure can be readily and rapidly (around three minutes for our dataset) applied to \texttt{aer2vec} representations that are generated without reference to predefined lexical knowledge. 

As with previous research \cite{aer2vec2019}, information in FAERS reports about PS designations proved to be very useful in improving model performance in all scenarios, including the \texttt{aer2vec} model, \texttt{retrofitting} algorithm, and even the PRR and ROR. One reason this field improves performance is that it designates reporters' suspicion and the reporters may have access to temporal information and background knowledge that is not explicit in the reports themselves. While there will likely be bias when expert opinion is involved, in this scenario, the trade-off appears warranted, to the extent our reference standards are reliable indicators of system accuracy after deployment. The effect of rescaling on \texttt{retrofitting} differs on two reference sets. As discussed above, rescaling increases AUCs by approximately 0.05 when evaluated on the OMOP set, but it has a negative effect with the EU evaluation set (a decrease about 0.02-0.03). Our working hypothesis is that the information encoded by magnitude is obscured in the case of the EU-FULL set, because the drugs in this set are far more frequently reported (median=18,457 vs. 2,395 for the OMOP set), and neural embeddings of words that occur in a broad variety of contexts tend to have shorter vectors due to representations of some of their surroundings canceling each other out \cite{schakel2015}. 

One concern about the current evaluation is that our training data (to 2019) extends past the point of publication of our evaluation sets (in 2013), which may mean that information that arose subsequently to the judgments in these sets was utilized. To assess whether or not this affected performance we re-ran our experiments using data from up to 2013 only, with $\alpha$ fixed at 0.5.  Relative performance of the methods was similar, with greatest improvements observed with retrofitting with rescaling in the context of the FULL set. Surprisingly, some experiments (e.g. with FULL models on the EU set) revealed better performance with this restricted data set, which suggests that in the absence of provider designation constraints these models may be disadvantaged by increases in false positive signals. We also evaluated performance across the four side effects in the OMOP set, and noted that performance benefits from the original retrofitting procedure were found in acute myocardial infarction and acute hepatic failure only. However with rescaling, performance for all four OMOP outcomes benefits from retrofitting. This explains the some of the improvement in  in overall performance, although further research is required to determine why some side effects benefit more from this procedure than others. 
% Also, the EU set is a smaller reference set, with a larger pool of adverse events, causing in general high variability in results with the FULL set. Consequently, the difference between normalization and rescaling is not significant (mean across all $\beta$, p=0.38, unpaired t-test), in contrast with results in the OMOP set (mean across all $\beta$, p=0.012, unpaired t-test).
\section*{Limitations}
While we have shown that performance measured against a widely-used reference set improves more with retrofitting, it is not yet fully apparent how this improvement is realized. While it seems self-evident that the retrofitted vectors incorporate information from drugs that are related in RxNorm and that this might be beneficial, in future work we plan to compare similarities between drug or ADE vectors of the same class before and after \texttt{retrofitting}. This becomes important when minimally preprocessed data are used. The DRUGNAME field contains exactly what was reported from a variety of sources without standardization. Synonyms or different spellings of a drug may disperse signal for drugs, which may fail to reach their generic names,as used in both the OMOP and EU evaluation sets. Future work will therefore involve closely monitoring how key drugs and their relatives are repositioned in vector space during the course of the retrofitting process. Also, only three relations (RO, RN, and SY) were used to build the lexicon. The explicit expansion of deeper relations (such as second-order ones) was also not implemented. Though one would anticipate the iterative nature of the retrofitting procedure capturing some of this information implicitly,  it remains possible that encoding the nature of these relationships explicitly would lead to further improvements in performance. An additional limitation is that the balance between distributional and lexical information is controlled by two hyper parameters ($\alpha$ and $\beta$) whose optimal values may depend on the original vectors, lexicons, and reference set. Furthermore, the two reference sets, EU and OMOP, were used for all evaluations.  As with previous work \cite{aer2vec2019} model performance on the OMOP set is generally lower than that on the EU set. One potential explanation for this concerns how the EU set is constructed, especially  its negative controls. According to the EU-ADR project \cite{euset}, Bayesian disproportionality analysis of the WHO spontaneous reporting database was used to create the pool for negative controls. This database is similar in nature to the FAERS, from which our training data were collected. In contrast, the OMOP project used product labeling, systematic literature review, and literature search to select  negative control candidates \cite{omop}. Lastly, these sets have been criticized for over-representation of well-established side-effects \cite{zoo}. This concern could be addressed by using time-delimited data, and reference examples from recently-added label warnings \cite{timeIndex}.

\section*{Conclusion}
We evaluated the applicability of the \texttt{aer2vec} algorithm to minimally preprocessed FAERS data, and the utility of  \texttt{retrofitting}, which augments \texttt{aer2vec} distributional representations with additional lexical information. \texttt{aer2vec} outperforms disproportionality metrics in most configurations, and the original implementation of \texttt{retrofitting} together with \texttt{aer2vec} further improves performance. Furthermore, our novel rescaled variant of the retrofitting algorithm results in significantly better performance on the larger (OMOP) reference set. The improvements in performance on a minimally preprocessed dataset make \texttt{retrofitting} a good alternative to explicit standardization, with potential application to a broad range of observational data sources.

\section*{Acknowledgements}
This work was supported by U.S. National Library of Medicine Grant (R01LM011563).

\makeatletter
\renewcommand{\@biblabel}[1]{\hfill #1.}
\renewcommand\refname{	
	\vspace{-2em}
	\begin{center}
		\text{References}
	\end{center}}
\makeatother

\bibliographystyle{unsrt}

\end{document}